# Cross-linguistic differences in gender congruency effects: Evidence from meta-analyses

Audrey Bürki[1], Emiel van den Hoven[1], Niels Schiller[2], & Nikolay, Dimitrov[1]

[1] Cognitive Science, Department of Linguistics, University of Potsdam,Karl-Liebknechstr. 24-25, 14476 Potsdam, Germany

[2] Faculty of Humanities, Leiden Univ. Centre for Linguistics, Reuvensplaats 3-4, 2311 BE Leiden, The Netherlands

Author note

The authors made the following contributions. Audrey Bürki: Conceptualization, Formal analyses, Supervision, Project administration, Writing - Original Draft Preparation, Writing - Review & Editing; Emiel van den Hoven: Conceptualization, Data collection & Pre-processing, Writing - Review & Editing; Niels Schiller: Writing - Review & Editing; Nikolay, Dimitrov: Formal analyses, Data collection & pre-processing, Writing - Original Draft Preparation.

Correspondence concerning this article should be addressed to Audrey Bürki, Karl-Liebknechstr. 24-25, 14476 Potsdam, Germany. E-mail: buerki@uni-potsdam.de

## Abstract

It has been proposed that the order in which words are prepared for production depends on the speaker's language. When producing the translation equivalent of *the small cat*, speakers of German or Dutch select the gender-marked determiner at a relatively early stage of production. Speakers of French or Italian postpone the encoding of a determiner or adjective until the phonological form of the noun is available. Hence, even though the words are produced in the same order (e.g., *die kleine Katze* in German, *le petit chat in French*), they are not planned in the same order and might require different amounts of advanced planning prior to production onset. This distinction between early and late selection languages was proposed to account for the observation that speakers of Germanic and Slavic languages, but not of Romance languages, are slower to name pictures in the context of a distractor word of a different gender. Meta-analyses are conducted to provide the first direct test of this cross-linguistic difference and to test a prediction of the late selection hypothesis. They confirm the existence of the gender congruency effect in German/Slavic languages and its absence in Romance languages when target and distractor words are presented simultaneously. They do not allow confirming the hypothesis that in the latter languages, a similar effect emerges when the presentation of the distractor is delayed. Overall, these analyses confirm the cross-linguistic difference but show that the evidence available to date is not sufficient to confirm or reject the late selection hypothesis as an explanation of this difference. We highlight specific directions for future research.

# Cross-linguistic differences in gender congruency effects: Evidence from meta-analyses

## Introduction

Thoughts are holistic (thinking about *a purple cow* does not consist of thinking of something indefinite, something purple, and of a cow, sequentially) but the use of language requires that these thoughts be organized in time, such that the speaker says “a purple cow”, one word after the other, and in this order. Models of sentence production generally assume that word forms are selected in the order of mention. In Meyer (1996; see also Jescheniak, Schriefers, and Hantsch, 2003) for instance, all words in a to be produced utterance are activated, with the amount of activation decreasing with the position of the word in the utterance. At the end of the 1990s, Miozzo, Caramazza, and colleagues formalized a novel hypothesis about determiner selection in the production of noun phrases (Costa, Sebastián-Gallés, Miozzo, & Caramazza, 1999; Miozzo & Caramazza, 1999; see also Alario & Caramazza, 2002; Caramazza, Miozzo, Costa, Schiller, & Alario, 2001; Miozzo, Costa, & Caramazza, 2002). According to this proposal, the order in which words are prepared for production does not necessarily follow the order of mention and may differ across languages. In “early selection languages”, speakers select the gender-marked determiner or adjective at a relatively early stage of production, namely as soon as the gender of the accompanying noun is retrieved. In “late selection languages”, phonological dependencies force speakers to postpone the selection of the determiner / gender-marked adjective until the phonological form of the noun is available. Examples of such dependencies are for instance found for singular definite determiners in French. The masculine and feminine definite determiners are respectively *le* and *la* before consonants (e.g., *la maison* “the house”; *le* *jardin* “the garden”) but *l’* before vowel-initial words, irrespective of their genders (e.g., *l’arbre* “the tree”).

This hypothesis (now commonly referred to as the “late selection hypothesis”) was proposed to account for a cross-linguistic difference in experimental effects. When producing a noun phrase with a gender-marked determiner (e.g., die Katze ‘the cat’) to describe a picture in the context of a word that is irrelevant to the task (i.e., distractor word), speakers of Germanic and Slavic languages are slower to name pictures in the context of a distractor word when the gender of the distractor and the gender of the target noun differ (see Schriefers, 1993 for the first demonstration, see also; La Heij, Mak, Sander, & Willeboordse, 1998; Schiller & Caramazza, 2003; Schriefers & Teruel, 2000; van Berkum, 1997). This gender congruency effect is however not observed when the determiner form is the same across genders (e.g., plural noun phrases in German, see Schiller & Caramazza, 2003) or when participants are asked to name the pictures with bare nouns (at least in Germanic languages, see Cubelli, Lotto, Paolieri, Girelli, and Job (2005) for reports of gender congruency effects with bare nouns in Italian and Spanish and Finocchiaro, Alario, Schiller, Costa, Miozzo, and Caramazza (2011) for null effects). The observation that gender congruency effects only surface when the phonological form of the determiner differs across genders led to the conclusion that these effects reflect competition between determiners (Caramazza et al., 2001). The exact level of processing at which this competition takes place is not clear. Whereas Caramazza et al. (2001) assume that the competition occurs during determiner form selection, other descriptions seem to imply that the competition arises during the selection of the semantic and grammatical features of the determiner (Miozzo et al., 2002). Crucially for our purposes, speakers of several Romance languages (e.g., Catalan, French, Italian, Spanish), on the other hand, seemed to be unaffected by the gender of the distractor when producing noun phrases with gender-marked determiners (Alario & Caramazza, 2002; Costa et al., 1999; Miozzo et al., 2002). Caramazza and collaborators related the absence of gender congruency effects in Romance languages to the presence of phonological dependencies between the gender marked determiner or adjective and the noun. In all the Romance languages tested at the time, the form of at least some determiners varied with the phonology of the next word. Later studies indeed suggested that in Romance languages without phonological dependencies between the determiner and the subsequent word, a gender congruency effect could be found (at least for subsets of nouns, Sá-Leite, Tomaz, et al., 2022). According to Caramazza et al.'s proposal, differences in gender congruency effects across languages reflect differences in the time course of determiner selection during utterance preparation. In the present study, we assess the evidence in favor of this proposal.

The theoretical relevance of the late selection hypothesis goes beyond the understanding of cross-linguistic differences. It speaks to the more general issue of gender representation and processing in language production. The mechanisms underlying gender selection in bare noun and utterance production are still a matter of debate in the psycholinguistic literature. The absence of gender congruency effects in Romance languages with dependencies together with the explanation of this finding in the context of the late selection hypothesis can contribute to this debate. Since the demonstration of gender congruency effects in Germanic/Slavic languages, many studies have been conducted to understand the mechanisms underlying these effects and their consequences for models of lexical access. These models assume that the preparation of a word for production starts with the activation of conceptual information. The conceptual layer then sends activation to lemmas, i.e., lexical representations specified for their semantic and grammatical properties (such as class or gender) in the next layer. Lemmas in turn send activation to the word form or phonological layer. According to a first view, gender congruency effects are associated with lemma access. Schriefers (1993) assumed for instance that each lemma is related to a gender node and that the congruency effect reflects competition between these nodes. In Roelofs (2018)'s influential model of word production Weaver++, the effect also originates at the lemma level but if the gender feature is always *activated*, it is only *selected* when it is required to determine the form of other words in the utterance. In addition, in this model, gender nodes do not compete, the first node to reach a pre-defined threshold is selected. These assumptions accommodate the findings that the gender congruency effect is absent in the production of bare nouns in Germanic languages (e.g., Finocchiaro et al., 2011; La Heij et al., 1998) or when determiner+noun utterances are produced with determiner forms that do not vary with gender (e.g., plural utterances in German or Dutch, see Schiller & Caramazza, 2003).

According to an alternative view, the gender congruency effect does not inform on gender feature selection for the noun but reflects the selection of determiner forms (Schiller & Caramazza, 2003). Accordingly, it is not a *gender* congruency effect but a *determiner* congruency effect. Sá-Leite et al. (2022) examined the evidence taken to support these different views with meta-analyses of gender congruency effects in all but gender-marked utterances in Romance languages (i.e., bare nouns in Romance languages, utterances with freestanding and bound morphemes with and without gender agreement in Germanic/Slavic languages). The only reliable finding was a gender congruency effect in gender-marked utterances with free-standing morphemes in Germanic/Slavic languages. The authors concluded that this finding (together with the absence of gender congruency effects for other utterance types) supports the hypothesis that the gender congruency effect is in fact a determiner congruency effect. We note, however, that the results of their meta-analysis are equally in line with the view that the effect reflects gender feature selection. The absence of significant meta-analytic estimates of gender congruency effects for some utterances, especially conducted in separate analyses, cannot be taken to support one or the other view. Crucially for our purposes, the absence of gender congruency effects in Romance languages as well as the late selection hypothesis as an explanation of this empirical result are difficult to reconcile with the view that gender congruency effects reflect the selection of the gender feature for the noun. Their investigation can therefore inform the locus of gender congruency effects in languages were these effects are observed.

We first focus on the claim that gender congruency effects are not found in languages with phonological dependencies, at least not in the experimental conditions in which they are observed in the Germanic/Slavic languages tested so far. Surprisingly, cross-linguistic differences in gender congruency effects are only supported by descriptive evidence. There is to date no direct comparison between languages. One study (Schriefers & Teruel, 1999) did address cross-linguistic differences but confounded language (German vs. French) with other relevant variables, such as word order, and did not report an analysis of the interaction between language and gender congruency. Other articles involved different languages (Costa et al., 1999; Miozzo et al., 2002) but never included languages with and without phonological dependencies, nor a statistical analysis of the interaction between language and gender congruency. The first aim of the present study is to obtain direct empirical evidence in support of the cross-linguistic difference in gender congruency effects. We take advantage of the many studies published on the gender congruency effect in the last three decades (see Sá-Leite et al., 2022 for a recent review). We report a meta-analysis of the gender congruency effect in each language type as well as a meta-regression to determine whether, across all published (and a couple of unpublished) studies, the evidence supports the hypothesis that the gender congruency effect is present in languages without phonological dependencies but absent in languages with such dependencies. In the last forty years, many studies were conducted to assess the processes and representations underlying word production, many of which used the picture-word interference paradigm. We now know that most these studies had low statistical power. Effects are generally small, sample sizes are small, variability across participants and items is high, with, as a consequence, a high probability of type I errors and a likely exaggeration of effect sizes in studies with significant effects. The absence of an effect in some studies and the presence of this same effect in other studies is therefore not sufficient to conclude that the effect is modulated by some properties of the design (e.g., language or relative timing of presentation of the distractor and picture). Likewise, results of single (underpowered) studies provide little information on the true size of an effect. Meta-analyses are the optimal tool to obtain this information retroactively as they summarize the information collected in past studies, taking into account the sample size and variability in each individual study.

Second, we assess the evidence in favour of the late selection hypothesis. Cross-linguistic differences in gender congruency effects do not inform on the time course of word selection in the languages being compared. Recall that the late selection hypothesis was proposed a posteriori, precisely to explain the absence of gender congruency effects in Romance languages with phonological dependencies. What would be needed is direct evidence that determiner selection in languages with dependencies occurs later than in Germanic or Slavic languages. A possible way to obtain this evidence is to investigate whether gender congruency effects in these languages surface when the competing determiner is activated later than in experiments with Germanic/Slavic languages. This can be achieved for instance by presenting the distractor at a later point in time, i.e., at later Stimulus Onset Asynchronies (or SOA). Several studies investigated gender conguency effects at later SOAs in Romance languages with dependencies, with mixed findings. Miozzo et al. (2002) did not find reliable evidence of a gender congruency effect at positive SOAs in Italian or Spanish. Foucart, Branigan, and Bard (2010) by contrast reported gender congruency effects when the distractor was presented 200ms after the picture (i.e., stimulus onset asynchrony or SOA of 200) in French. In our own work in the same language, we were not able to replicate this finding but found that gender congruency effects can be detected when the model adjusts for inter- and intra-individual differences in the time required to process pictures and distractors, even at an SOA of 0ms (simultaneous presentation of picture and distractor word, Bürki, Besana, Degiorgi, Gilbert, & Alario, 2019). The second aim of the present study is to seek evidence in favour of the late selection hypothesis by examining the prediction that gender congruency effects can be found in Romance languages with phonological dependencies at later SOAs. To do so, we again rely on a meta-analysis of previously published work on this issue.

## Method

### Literature review, eligibility criteria and dataset selection

Our search for relevant studies was conducted as follows. We first extracted all potentially relevant references from a set of articles on the gender congruency effect that we already had. We searched for articles citing the seminal articles by Schriefers (1993) and Miozzo and Caramazza (1999). We also searched several databases (www.proquest.com; www.scopus.com; https://apps.webofknowledge.com) using the queries "determiner congruency", "determiner-congruency", "gender congruency", "gender-congruency".

Experiments were included if they fulfilled all the following criteria. The participants were healthy young adults, native speakers of the language used in the study. The language of the study was either a Germanic, Slavic, or Romance language with phonological dependencies. The task was a classical picture-word interference task: a picture was presented on the screen accompanied by a distractor. The task of the participants was to name the picture using a multi-word utterance made of a gender-marked freestanding morpheme followed by a noun. Moreover, in Romance languages, the phonological form of the freestanding morpheme (at least for one gender) depended on the onset of the next word. In order to assess the evidence supporting gender congruency effects at early SOAs in Germanic/Slavic languages and study potential differences in these effects in Romance languages with dependencies at these SOAs, we selected all experiments or experiment parts in which a written or spoken noun (the distractor) and the target word were presented simultaneously (SOA = 0). In order to assess the evidence supporting a gender congruency effect at later SOAs in Romance languages with dependencies, we further selected, for these languages, experiments or experiment parts in which the distractor was presented 150ms, 200ms, or 300ms after picture onset (thereafter "late SOAs"). We considered (somewhat arbitrarily) that an SOA of 100ms was not long enough and therefore disregarded the three experiments with this SOA along with one experiment with a negative SOA.

We contacted all of the authors of the studies that are included in the paper asking for the raw data and any unpublished data, though not all of them replied. We also included one of our own unpublished datasets. Several of the experiments relevant for the meta-analysis manipulated additional factors (e.g., semantic or phonological relatedness between target and distractor words, distractor frequency). When these manipulations were orthogonal to the gender congruency manipulations, we ignored them.

At the end of this procedure, we were left with 41 experiments from 19 different articles. We split experiments in separate studies when more than one SOA was tested, which resulted in 53 datasets. This was necessary to assess the impact of SOA on the gender congruency manipulation.

**Data availability statement**

The scripts and datasets to reproduce the analyses and the present paper can be found on OSF (https://osf.io/g6vax/?view_only=5019d6555df54e5197a1d348c24ffac5).

**Extraction of estimates of gender congruency effects in individual studies**

The input to the meta-analysis is an estimate of the effect and its standard error in individual studies. When the raw dataset was available (8), we computed these values by fitting a linear mixed-effects model with untransformed speech onset latencies as the dependent variable and the predictor gender congruency as fixed effect. Gender-congruent trials were coded -0.5 and gender-incongruent 0.5. This way, a positive estimate means that gender-incongruent items were produced slower than gender-congruent items. Each model included all the random intercepts and slopes allowed by the design. Note that this analysis may differ from the analysis originally performed by the authors of the studies and as a consequence, the estimates may show slight differences. When the raw dataset was not available for a given study, we extracted or computed the effect size and its standard error based on the information in the published paper (23). For those cases where there was no information on the standard error (the authors reported the comparison of interest to be non-significant, or $F < 1$), we used the mean standard error across available datasets as an estimate of the standard error for these studies (22) .

**Analysis**

Table 1:

*List of meta-analyses and meta-regressions*

| Analysis | Languages | N | SOA | Effect |
|---|---|---|---|---|
| Meta-analysis | Germanic and Slavic | 22 | early | Congruency |
| Meta-analysis | Romance with dependencies | 21 | early | Congruency |
| Meta-regression | All | 43 | early | Language |
| Meta-analysis | Romance with dependencies | 12 | late | Congruency |
| Meta-regression | Romance with dependencies | 33 | early-late | SOA |

Table 1 lists the meta-analyses and meta-regressions conducted for the purpose of the present study. Meta-analyses provide an estimate of the size of the gender congruency effect and of its degree of uncertainty, based on the effect size and standard error in a specific set of studies. In meta-regressions, the gender congruency effect is estimated for different levels of a predictor, i.e., language type or SOA.

The aim of the first analysis was to determine the size of the gender congruency effect in German and Slavic languages. The aim of the second and third analyses was to seek evidence in favor of the claim that the gender congruency effect is present in these languages but absent in Romance languages with dependencies. The remaining analyses were run on Romance languages with phonological dependencies and aimed at quantifying the effect at late positive SOAs as well as the difference in the effect size between late SOAs and SOAs of 0ms.

All analyses were performed in the Bayesian framework, in R (R Core Team, 2022). Bayesian analyses provide information on the probability of an effect given the data and the model. The output of the analysis is a distribution of all possible values of the estimated effect. In the present study, and following standard practice, we report the 95% credible interval, which tells us the range within which the true value of the effect has 95% probability of falling. We further report Bayes Factors to quantify the evidence in favor of the null hypothesis of no effect over the alternative hypothesis that there is a gender congruency effect (or a modulation of this effect by language type or SOA). We use Lee and Wagenmakers (2014)'s version of Jeffreys (1998)'s scale to interpret Bayes Factors and considered values between 1 and 3 as anecdotal evidence, values between 3 and 10 as moderate evidence, and values between 10 and 30 to provide strong evidence. In the computation of Bayes Factors, Null models always consisted of models with the same random effects structure as the alternative model. To calculate Bayes factors we used bridge sampling (Bennett, 1976; Gronau et al., 2017).

Meta-analyses in the Bayesian framework take as input the size of the effect of interest (i.e., the difference between the mean for incongruent trials and the mean for congruent trials) and the standard error of that estimate in individual studies as well as prior distributions for each of the model parameters. These priors must be defined by the researcher. They represent a priori assumptions about the size of the effect of interest. Following standard practice, we used weakly informative priors to estimate effect sizes and credible intervals, and slightly more constraining priors to compute Bayes Factors (this is because weakly informative priors tend to favor the null hypothesis). Moreover, in each case, we performed a sensitivity analysis, that is, we reproduced the same analysis with different priors to get a sense of the influence of the priors on the results. Only the priors for the estimate of interest (intercept in meta-analyses and intercept and slopes in meta-regressions) were varied across analyses. For all other parameters (i.e., standard deviations), we used $\mathcal{N}_{+}(0,100)$ (truncated normal).

Models for meta-analyses only had an intercept and a random intercept for the random factor *study*. Models for meta-regressions had an intercept, a random factor for study, as well as a slope for the predictor. Whenever we report the meta-analytic estimate or estimate for the predictor and their 95% credible intervals in the text or graphs, the following prior was used for the intercept / slope : $\mathcal{N}(0,100)$. This prior reflects the assumption that the value for the parameter lies with 95% probability between -200 and 200ms, with a higher probability for values around 0. In the sensitivity analysis for the estimates, we report the output of models with two additional priors for the same parameters, namely $\mathcal{N}(0,200)$ and a uniform distribution bounded between -100 and 100.

For Bayes factors, we used the following range of more informative priors for the intercept and the slope : $\mathcal{N}(0,10)$, $\mathcal{N}(0,20)$, $\mathcal{N}(0,40)$, $\mathcal{N}(0,80)$. All these priors assume that the most likely value is 0 and are agnostic with regard to the direction of the effect. They differ in the range of plausible values. A prior of $\mathcal{N}(0,10)$ assumes that the effect has a 95% probability of lying between -20 and 20 ms, with values around 0 being more likely. A prior of $\mathcal{N}(0,80)$ assumes that the effect has a 95% probability of lying between -160 and 160 ms.

We opted for random-effects rather than for fixed-effects meta-analyses/meta-regressions. Random-effects meta-analyses/regressions assume that each study has a different true effect $\theta_i$. Given that the individual studies were conducted in different labs, in different languages and with different materials, we assumed that the effect would not be identical across studies.

All meta-analyses were conducted under the following set of assumptions. Each study $i$ out of a total of $n$ studies included in the meta-analysis has an underlying true gender congruency effect $\theta_i$. This true effect is assumed to come from a normal distribution with mean $\theta$ and standard deviation $\tau$. The observed effect $y_i$ in each study $i$ is assumed to stem from a normal distribution with mean $\theta_i$ and standard deviation $\sigma_i$ (true standard error of the effect in the study). The specifications of the meta-analytic models (assuming $\mathcal{N}(0,100)$ for the prior distribution of the effect size) are displayed in Equation (1) where $\theta$ is the true gender congruency effect to be estimated by the model, $y_i$ is the observed effect in a given study; $\sigma_i$ is the standard error for study i, and $\tau$ is the between-study standard deviation.

$$\begin{aligned} y_i|\theta_i, \sigma_i &\sim \mathcal{N}(\theta_i, \sigma_i), i = 1, \dots, n \\ \theta_i|\theta, \tau &\sim \mathcal{N}(\theta, \tau), \\ \theta &\sim \mathcal{N}(0,100), \\ \tau &\sim \mathcal{N}_+(0,100) \end{aligned} \tag{1}$$

For meta-regressions, there is an additional parameter to be estimated by the model, the regression coefficient $\beta$:

$$\begin{aligned} y_i|\theta_i, \beta, \sigma_i^2 &\sim \mathcal{N}(\theta_i + \beta * predictor_i, \sigma_i^2), i = 1, \dots, n \\ \beta &\sim \mathcal{N}(0,100) \end{aligned} \tag{2}$$

**Results**

***Gender congruency at SOA = 0 in Germanic/Slavic languages and Romance languages with phonological dependencies.*** Figure 1 shows the posterior distributions of the gender congruency estimates for each study as well as the meta-analytic estimates and 95% credible intervals for German/Slavic languages as well as Romance languages with phonological dependencies. Figure 2 displays the posterior distribution of the meta-analytic estimate of the gender congruency effect evaluated separately for Romance and Slavic/Germanic languages as well as the corresponding Bayes factors. Sensitivity analyses can be found in Appendix B. The results of the meta-regression testing for the effect of language type on the gender congruency effect are illustrated in Figure 3.

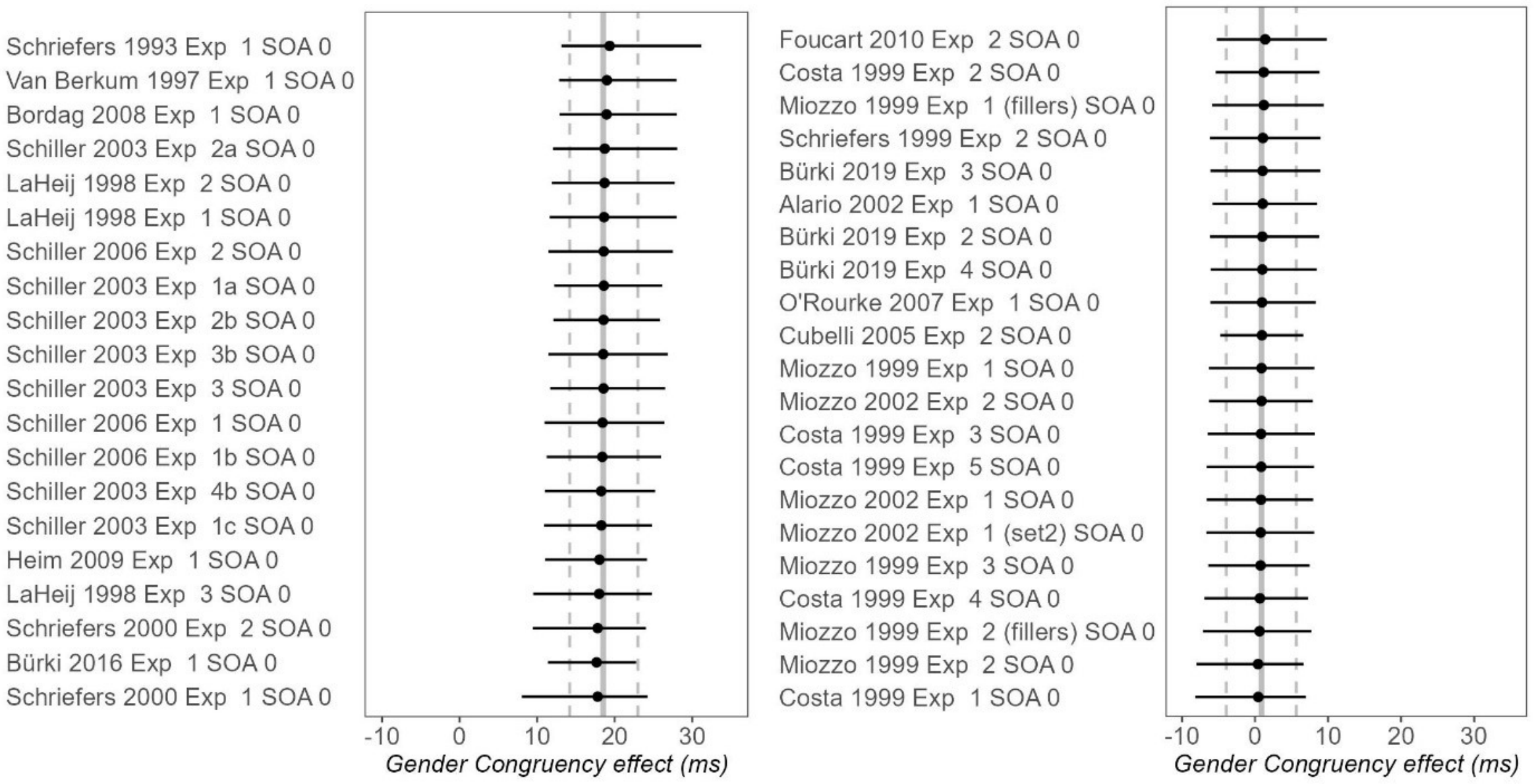

*Figure 1.* Summary of the random-effects meta-analyses modeling the effect of gender congruency on naming times for Germanic/Slavic languages (left) and Romance languages with phonological dependencies (right). For each study, the figure displays, in black, the mean and posterior estimate (mean and 95% credible interval). A positive value means that target-distractors pairs with incongruent genders result in longer naming latencies for the picture. The grey vertical line represents the grand mean (i.e., the meta-analytic effect) and the dashed vertical lines delimit the 95% credible interval of that estimate.

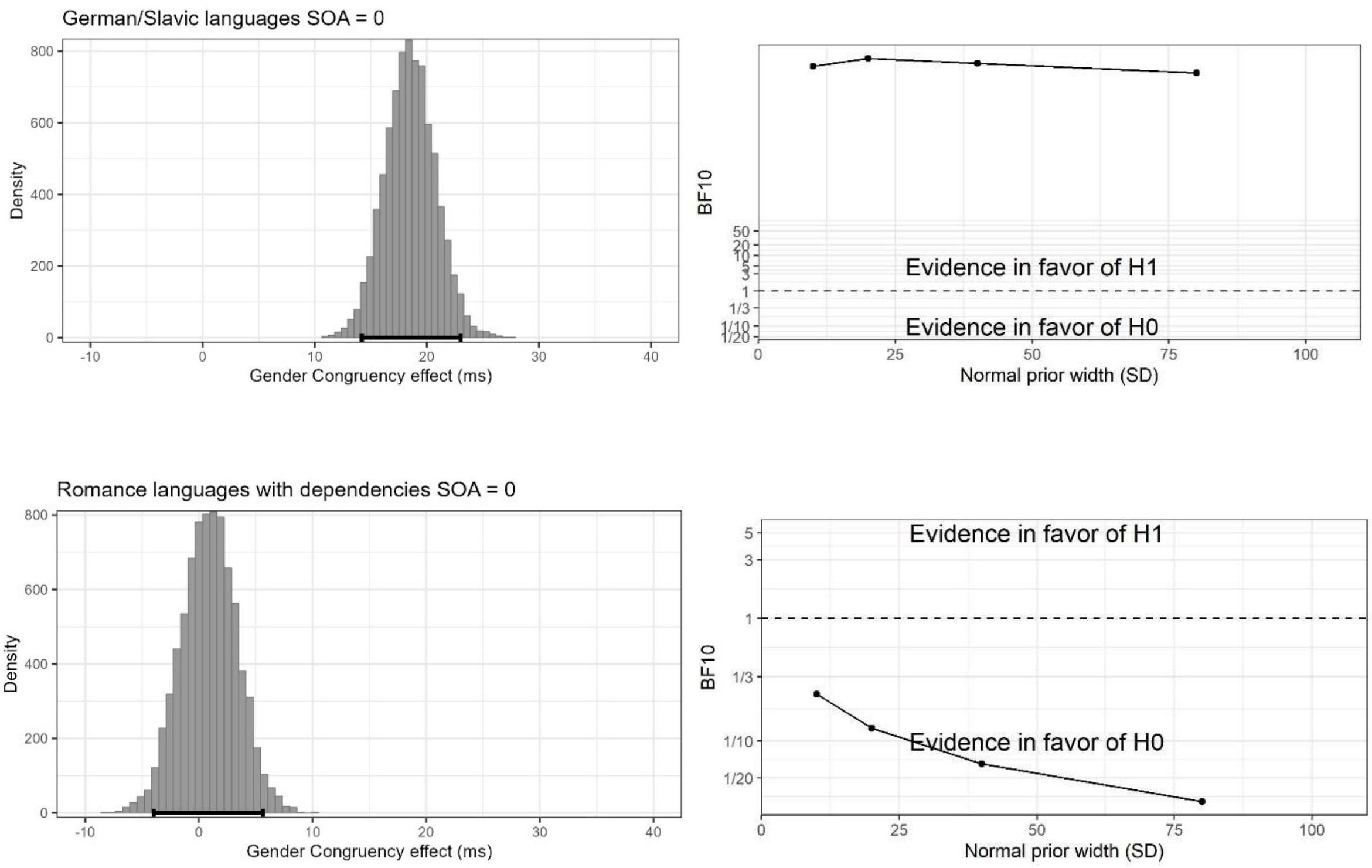

*Figure 2.* Results of meta-analyses of gender congruency effect for Germanic/Slavic languages (upper panels) and for Romance languages with phonological dependencies (lower panels) at SOA = 0 (Posterior distributions with 95% CrI and Bayes Factors with different priors).

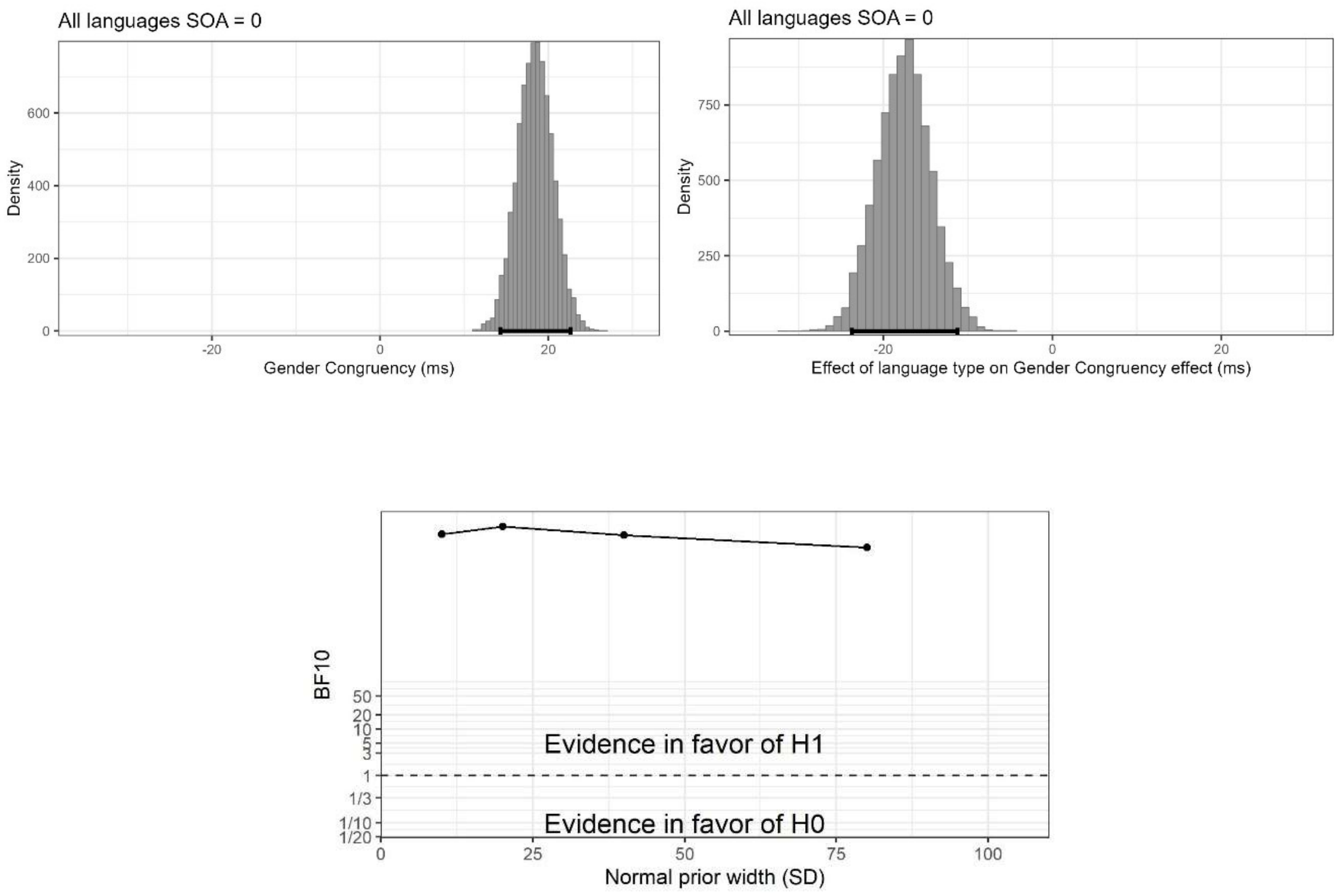

*Figure 3.* Posterior distributions (with 95% CrI) and Bayes Factors for the meta-regression evaluating the impact of language type on the gender congruency effect

These analyses show that when considering all published studies with German and Slavic languages in which the picture and distractor are presented simultaneously (n = 20), the overall effect of gender congruency is 18.52ms [14.20, 23.00]. We note that these values are similar to the values reported in Bürki et al. (2020)'s meta-analysis of the semantic interference effect (longer naming times in bare picture naming when the distractor word is a member of the same semantic category than when it is unrelated to the to-be-produced word) which was about 21 ms (95% Credible interval 18–24 ms) using a similar statistical approach. By contrast, in Romance languages with phonological dependencies at this same SOA, the effect is 0.89ms, [-3.95, 5.65]. Bayes Factors provide strong evidence that the effect differs from zero in Germanic/Slavic languages and moderate to strong evidence that the effect does not differ from zero in the analyzed Romance languages. The results of the meta-regression confirm this pattern and provide strong evidence in favor of the hypothesis that at SOA 0, the gender congruency effect depends on whether the language investigated has or does not have phonological dependencies. These analyses provide the first direct evidence that the gender congruency effect depends on the language being studied.

The dominant hypothesis to explain this discrepancy assumes that determiners are selected later in Romance languages with dependencies. Finding a gender congruency effect when the presentation of the distractor is delayed such that the determiner form associated with the distractor can compete with the determiner form of the target noun would provide support for this hypothesis. This is examined in the next set of analyses.

***Gender congruency at early (0ms) vs. late SOAs in Romance languages with phonological dependencies***. Figure 4 shows the posterior distribution of the estimate of the gender congruency effect for each study included in this meta-analysis. Figure 5 displays the posterior distribution of the effect of gender congruency in Romance languages with phonological dependencies at late SOAs (150-300ms) and the corresponding Bayes factors with different priors. Sensitivity analyses can be found in Appendix C. The results of the meta-regression testing for the influence of SOA on the gender congruency effect in Romance languages are illustrated in Figure 6. In this analysis, contrasts were set for the variable SOA such that late SOAs have a value of 0.5 and SOAs of 0 a value of -0.5.

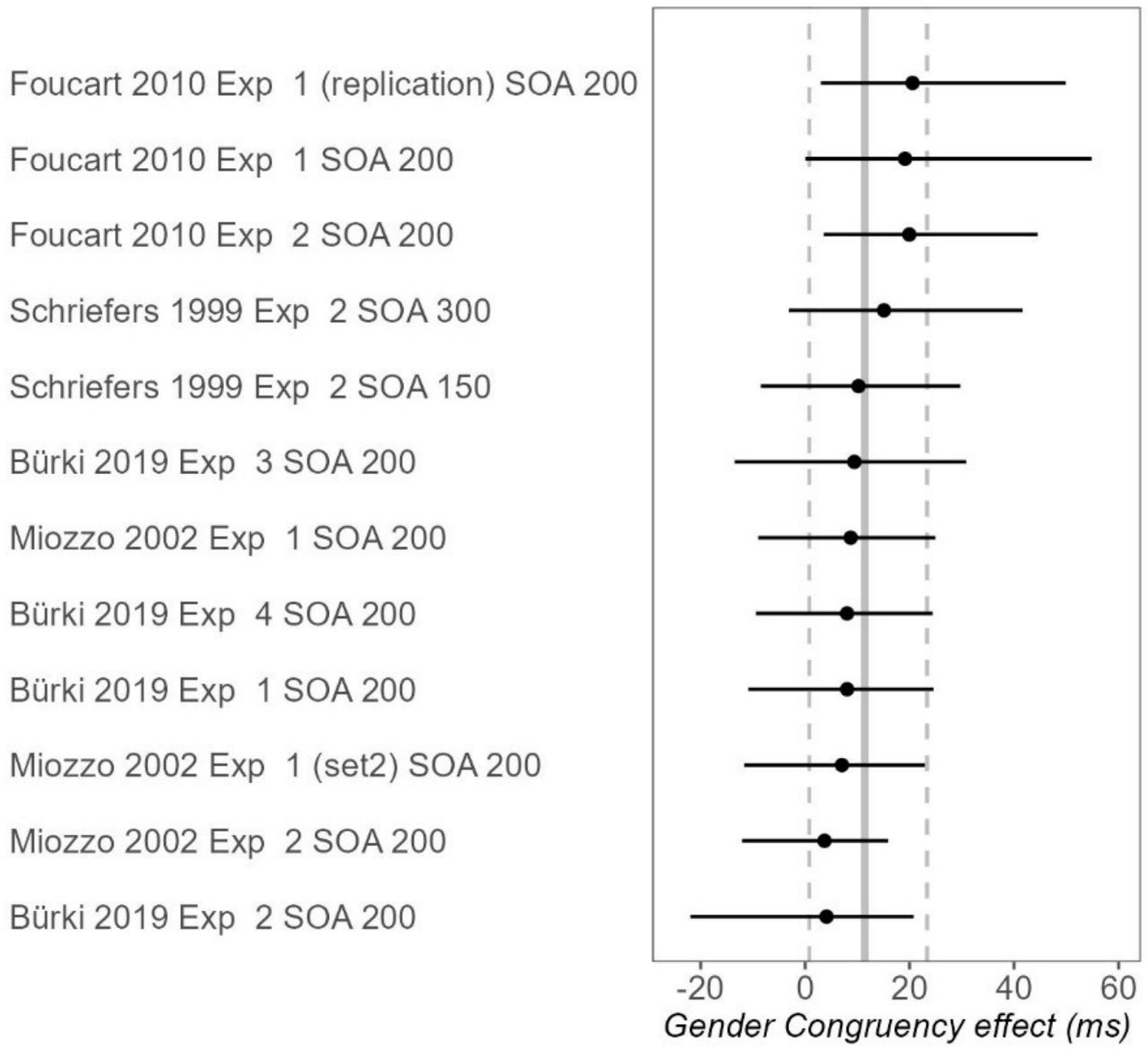

*Figure 4.* Summary of the random-effects meta-analysis modeling the effect of gender congruency on naming times for Romance languages with phonological dependencies at positive SOAs. For each study, the figure displays, in black, the mean and posterior estimate (mean and 95% credible interval). A positive value means that target-distractors pairs with incongruent genders result in longer naming latencies for the picture. The grey vertical line represents the grand mean (i.e., the meta-analytic effect) and the dashed vertical lines delimit the 95% credible interval of that estimate.

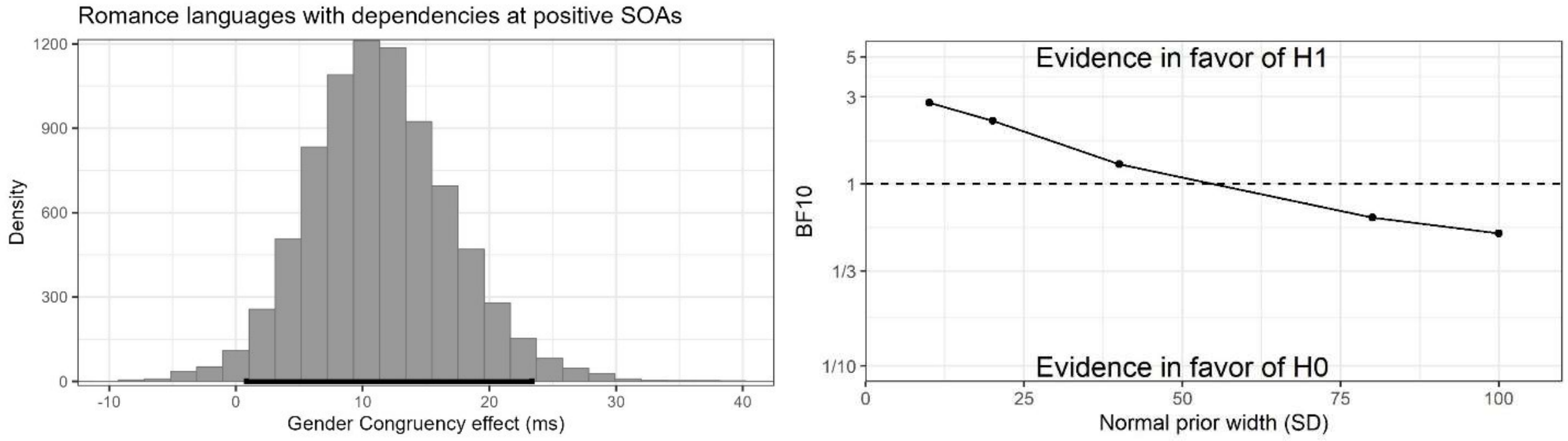

*Figure 5.* Posterior distribution (with 95% CrI) and Bayes Factors for the meta-analysis of the gender congruency effect for Romance languages with phonological dependencies at positive SOAs

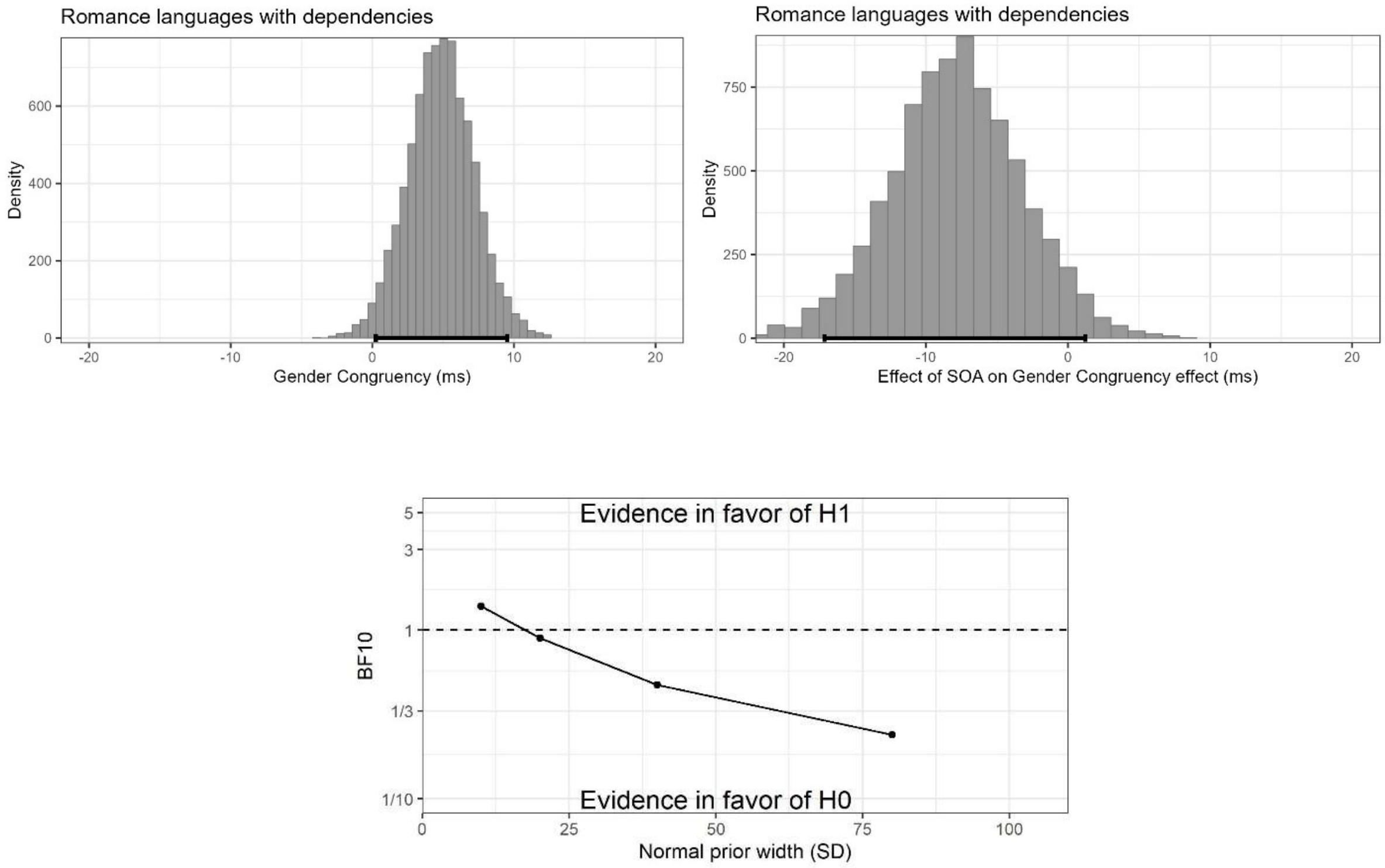

*Figure 6.* Posterior distributions (with 95% CrI) and Bayes Factors for the meta-regression evaluating the impact of SOA on the gender congruency effect for Romance languages with phonological dependencies

This second set of analyses shows that when considering published studies on Romance languages with phonological dependencies in which the distractor was presented 150 to 300ms after picture onset, there is no evidence in favor or against the hypothesis that the congruency in genders between target and distractor words impacts naming latencies. The meta-analytic estimate is in the expected direction, but the 95% credible interval is large, signalling high uncertainty in the estimate (11.42ms [0.82, 23.35]). More data seem necessary to obtain a more precise estimation. Twelve studies are included in this analysis (20 were included in the analysis with Germanic/Slavic languages, with as a result, a higher precision). Following the request of a reviewer, we ran the same analysis, this time including the three studies with an SOA of +100ms. The meta-analytic estimate is slightly lower (8.08ms) and the 95% credible contains zero ([-0.22, 17.72]). Bayes Factors are inconclusive, except the one with the largest prior, which shows moderate evidence in favor of the hypothesis of no effect.

For completeness, we also ran a meta-analysis with the five studies with SOAs ranging from 150 to 300 in German/Slavic languages. The gender congruency effect is 6.78ms, with a large credible interval that includes 0 ([-8.77, 22.40]). Bayes factors provide no evidence supporting either the hypothesis that there is an effect at these SOAs or that there is no effect. The meta-regression testing for the impact of SOA on the gender congruency effect provides anecdotal to moderate evidence (depending on the priors) in favor of the hypothesis that SOA does not impact the size of the gender congruency effect in Germanic/Slavic languages.

## General Discussion

Cross-linguistic differences in gender congruency effects have had a substantial influence on theorizing about the processes and representations underlying language production. According to Caramazza et al. (2001)'s proposal, languages differ with regard to the order in which words are prepared for production. Moreover, this order does not necessarily correspond to the order of mention. This proposal did not only contribute a new theoretical perspective, it further pointed to possible ways of tracing the time course of word selection/encoding during the preparation of an utterance and their differences across languages. Despite its important implications, this proposal was not supported by statistical analyses, i.e., independent studies reported null effects in Romance languages with dependencies and significant effects in Germanic/Slavic languages. Our meta-analyses/regressions provide the first direct statistical evidence supporting a cross-linguistic difference in gender congruency effects at SOAs of 0ms. Considering the available evidence, our analyses show that when target pictures and distractors are presented jointly, the gender congruency effect is present in Germanic/Slavic languages but absent in Romance languages with dependencies. This intriguing cross-linguistic difference calls for an explanation.

The late selection hypothesis was proposed a posteriori, to explain the (at the time anecdotal) evidence for a difference in gender congruency effects across languages. This difference can thus hardly be taken to support the late selection hypothesis. The observation of gender congruency effects when the activation of the determiner associated with the distractor is delayed would provide direct empirical support in favor of the late selection hypothesis, and the more general claim that the order in which words are prepared for production differs across languages. We conducted a meta-analysis with 12 studies on Romance languages with phonological dependencies in which the distractor was presented 150 to 300ms after picture onset to assess the evidence in favour of gender congruency effects at these SOAs. Interestingly, the estimates for individual studies, weighted by the meta-analysis, all show a descriptive effect in the expected direction, that is, longer response times for incongruent trials. The mean estimate is positive but the standard error is large, which prevents any conclusion.

As mentioned in the Introduction, the theoretical relevance of the late selection hypothesis goes beyond the understanding of cross-linguistic differences. The late selection hypothesis has contributed to debates about the locus of gender congruency effects, in languages and utterances in which these effects are observed. This hypothesis indeed implies that gender congruency effects reflect determiner selection (Caramazza et al., 2001). In the light of the present study's findings, and until further evidence is provided, the strength of this argument needs to be reconsidered. In their meta-analyses of gender congruency effects in bare nouns (in Romance languages) and in utterances with free-standing and bound morphemes (in Germanic/Slavic languages), Sá-Leite et al. (2022) showed that the only reliable effect is found for gender-agreeing utterances with free-standing morphemes in Germanic/Slavic languages. They concluded that this finding is consistent with a *determiner* congruency effect. In our opinion, however, this finding alone does not inform on the locus of the effect and is also compatible with the view that the effect reflects gender feature selection. The absence of significant meta-analytic estimates of gender congruency effects for other types of utterances (especially conducted in separate analyses) cannot be taken to support one or the other view. Taken together, the findings of Sá-Leite et al. (2022) and that of the present study, which probably cover most of the studies manipulating gender congruency between target and distractor in a speaker's native language thus far, do not allow clear conclusions regarding the mechanisms underlying gender congruency effects and thereby gender processing and representation during language production. In the remainder of this General Discussion, we discuss potential ways forward.

The present study confirms the cross-linguistic difference in gender congruency effects at early SOAs and highlights the lack of support for the late selection hypothesis, an intuitive explanation of this cross-linguistic difference. Additional studies are clearly needed to either test alternative explanations of the absence of gender congruency effects in some languages or to gather additional data in favor (or against) the late selection hypothesis. Notably, the late selection hypothesis does not specify the SOA at which a gender congruency effect could occur in Romance languages with dependencies. The studies included in the present analysis tested SOAs of +150ms, +200ms or +300ms (this latter SOA was tested in one study only). Gender congruency effects could however surface at later SOAs and an important goal for future studies on this issue will be to examine a wider range of SOAs. If gender congruency effects do surface at later SOAs in Romance languages with phonological dependencies, however, picking up on these effects with classical analyses is likely to be harder than in early selection languages. According to the late selection hypothesis, determiner selection in Romance languages with dependencies takes place once the phonological form of the noun (or at least the first phoneme) has been accessed. The speed with which word forms are accessed e.g., in picture naming tasks, is known to vary greatly across items. The frequency of the noun or the age at which it was acquired for instance impact naming times (e.g., Alario et al., 2004; Ellis & Morrison, 1998; Jescheniak & Levelt, 1994). In addition, participants differ in the time they need to prepare words for production (Laganaro, Valente, & Perret, 2012; Shao, Roelofs, & Meyer, 2012). Irrespective of whether this variability originates during the access to word forms or during the processes taking place before that, it impacts the time at which word forms are accessed. It follows that the optimal SOA for the gender congruency effect to surface is likely to vary across items and participants. The observed larger standard error for the estimate of the gender congruency effect at late SOAs in the analysed Romance languages is as expected under the hypothesis that the effect is indeed more variable across studies in these languages at this SOA. Likewise, a close look at Figures 1 and 4 suggests that not only the meta-analytic estimate but also the estimates of individual studies are more variable at late SOAs in Romance languages with phonological dependencies. In the absence of direct comparisons, greater variability in a subset of languages or SOAs should however be interpreted with caution (the datasets differ in sample sizes as well as in the information used to compute the standard errors in individual studies). Further investigations of the late selection hypothesis might want to compare not only the mean effect size across languages and SOAs but also the variability around this estimate. In addition, these studies will need to perform systematic comparisons with wider ranges of SOAs, and to compare languages with similar material (e.g., target words and distractor words with similar lexical properties across languages, target-distractor pairs with the same types of semantic relationships across languages). Existing data mostly come from experiments in which different languages were examined in different studies, using different material.

Another potential way forward might be to implement the difference in timing between distractor and target word at the trial level rather than with SOAs, by registering, in independent tasks and for each target and distractor word, the difference between the time needed to produce the target word (without a distractor) and to process the distractor alone. In previous work, we found that other experimental effects with the same paradigm (phonological facilitation, semantic interference, Bürki & Madec, 2022) do interact with a difference measure of this kind. Our attempt to find a similar interaction with the gender congruency effect in French has failed (Bürki et al., 2019) but we found that accounting for the difference in processing times between the two words in the statistical model did increase the size of the gender congruency effect.

Whereas systematic investigations of gender congruency effects at later SOAs might prove useful to further assess the late selection hypothesis, efforts should also be made to find alternative paradigms. Importantly, a convincing demonstration that gender congruency effects do not occur in Romance languages with dependencies would not necessarily speak against the late selection hypothesis. It could still be argued that the absence of gender congruency effects at early SOAs in these languages is indeed due to a later selection of the determiner, but that this does not translate into a gender congruency effect (that is, a difference in naming times between congruent and incongruent trials) at any SOA. Miozzo et al. (2002) suggested for instance that interference from a competing determiner may occur in Romance languages with dependencies when distractor and picture are presented simultaneously but that this interference could only impact response latencies "if the time to resolve the interference produced by the gender-congruent distractors exceeded the time necessary for acquiring the critical information from the phonological phrase." (p. 389).

The present study complements recent attempts to establish the strength and reliability of experimental effects in the picture-word interference paradigm using meta-analyses. The general picture seems to be that experimental effects such as semantic interference (Bürki et al., 2020b), distractor frequency (Bürki et al., 2020a) or gender congruency (Sá-Leite et al., 2022) are reliable effects. Importantly however, the evidence regarding the variables that modulate these effects, which is often necessary to disentangle opposing accounts of these effects, is much less convincing. There is an urgent need for the field to undertake systematic investigations of these moderators and to find alternative paradigms.

Before we conclude we would like to highlight two important methodological contributions of the present study. The first meta-analysis provides an estimate of the size of the gender congruency effect which can be used for two purposes. First, meta-analytic estimates can be used as upper bounds to guide sample size selection in future studies. We note that this effect is small. As a consequence, large sample sizes are required to ensure sufficient power. If we assume an effect of the same size for Romance languages with phonological dependencies at later SOAs but with additional variability, detecting this effect will require even larger sample sizes.

Second, current models of determiner selection / time course of word selection are purely theoretical. A computational implementation would help generating specific predictions with regard for instance to the range of SOAs at which an effect could be observed under the late selection hypothesis. The meta-analytic estimate can be used to inform such computational models.

To summarize, the present study confirms cross-linguistic differences in the gender congruency effect during the production of utterances with freestanding morphemes. This finding provides the field with a rare opportunity to dig into cross-linguistic differences in processing during utterance preparation. The present study further highlights the lack of independent evidence in favor of the late selection hypothesis as an explanation of this difference and calls for additional data and the consideration of alternative explanations. This work further highlights several avenues to clarify these issues in future work and provides estimates to guide future experimental studies and computational modeling.

## Acknowledgments

This research was funded by the Deutsche Forschungsgemeinschaft (DFG, German Research Foundation), project numbers BU 3542/2-1 and 317633480 (in SFB 1287, Project B05, PI : Audrey Bürki).

**Appendix 1**

*Datasets included in the meta-analyses and their properties*

| Dataset | Language.Type | Language | SOA | Participants.N. | Pictures.N. | SE |
|---|---|---|---|---|---|---|
| Schriefers 1993 Exp. 1 | No dependencies | Dutch | 0 | 18 | 10 | Paper |
| Van Berkum 1997 Exp. 1 | No dependencies | Dutch | 0 | 24 | 24 | Paper |
| LaHeij 1998 Exp. 1 | No dependencies | Dutch | 0 | 20 | 16 | Average |
| LaHeij 1998 Exp. 2 | No dependencies | Dutch | 0 | 20 | 16 | Average |
| LaHeij 1998 Exp. 3 | No dependencies | Dutch | 0 | 16 | 16 | Average |
| Costa 1999 Exp. 1 | Dependencies | Catalan | 0 | 20 | 50 | Paper |
| Costa 1999 Exp. 2 | Dependencies | Catalan | 0 | 20 | 20 | Paper |
| Costa 1999 Exp. 3 | Dependencies | Spanish | 0 | 21 | 20 | Average |
| Costa 1999 Exp. 4 | Dependencies | Spanish | 0 | 20 | 50 | Average |
| Costa 1999 Exp. 5 | Dependencies | Spanish | 0 | 21 | 50 | Average |
| Miozzo 1999 Exp. 1 (fillers) | Dependencies | Italian | 0 | 12 | 30 | Paper |
| Miozzo 1999 Exp. 1 | Dependencies | Italian | 0 | 12 | 22 | Average |
| Miozzo 1999 Exp. 2 (fillers) | Dependencies | Italian | 0 | 15 | 30 | Average |
| Miozzo 1999 Exp. | Dependencies | Italian | 0 | 15 | 22 | Paper |
| 2 | | | | | | |
| Miozzo 1999 Exp. 3 | Dependencies | Italian | 0 | 25 | 27 | Average |
| Schriefers 2000 Exp. 1 | No dependencies | German | 0 | 16 | 18 | Average |
| Schriefers 2000 Exp. 2 | No dependencies | German | 0 | 16 | 18 | Paper |
| Miozzo 2002 Exp. 1 | Dependencies | Italian | 0 | 15 | 22 | Average |
| Miozzo 2002 Exp. 1 | Dependencies | Italian | 200 | 15 | 22 | Average |
| Miozzo 2002 Exp. 1 (set2) | Dependencies | Italian | 0 | 15 | 30 | Average |
| Miozzo 2002 Exp. 1 (set2) | Dependencies | Italian | 200 | 15 | 30 | Average |
| Miozzo 2002 Exp. 2 | Dependencies | Spanish | 0 | 18 | 50 | Average |
| Miozzo 2002 Exp. 2 | Dependencies | Spanish | 200 | 18 | 50 | Average |
| Schiller 2003 Exp. 3 | No dependencies | Dutch | 0 | 26 | 22 | Paper |
| Schiller 2003 Exp. 3b | No dependencies | Dutch | 0 | 26 | 22 | Paper |
| Schiller 2003 Exp. 1a | No dependencies | German | 0 | 27 | 60 | Paper |
| Schiller 2003 Exp. 1c | No dependencies | German | 0 | 26 | 60 | Paper |
| Schiller 2003 Exp. 2a | No dependencies | Dutch | 0 | 17 | 22 | Paper |
| Schiller 2003 Exp. 2b | No dependencies | Dutch | 0 | 18 | 22 | Paper |
| Schiller 2003 Exp. 4b | No dependencies | Dutch | 0 | 15 | 22 | Paper |
| Cubelli 2005 Exp. 2 | Dependencies | Italian | 0 | 28 | 16 | Paper |
| Schiller and Caramazza 2006 Exp. 1 | No dependencies | Dutch | 0 | 28 | 24 | Paper |
| Schiller and Costa 2006 Exp. 1b | No dependencies | German | 0 | 20 | 28 | Paper |
| Schiller and Caramazza 2006 Exp. 2 | No dependencies | Dutch | 0 | 19 | 24 | Paper |
| Bordag 2008 Exp. 1 | No dependencies | Czech | 0 | 32 | 36 | Paper |
| Heim 2009 Exp. 1 | No dependencies | German | 0 | 14 | 60 | Paper |
| Foucart 2010 Exp. 1 | Dependencies | French | 200 | 18 | 48 | Paper |
| Foucart 2010 Exp. 2 | Dependencies | French | 0 | 18 | 48 | Average |
| Foucart 2010 Exp. 2 | Dependencies | French | 200 | 18 | 48 | Average |
| Foucart 2010 Exp. 1 (replication) | Dependencies | French | 200 | 18 | 48 | Paper |
| Schriefers 1999 Exp. 2 | Dependencies | French | 0 | 16 | 8 | Average |
| Schriefers 1999 | Dependencies | French | 150 | 16 | 8 | Average |
| Exp. 2 | | | | | | |
| Schriefers 1999 Exp. 2 | Dependencies | French | 300 | 16 | 8 | Paper |
| O'Rourke 2007 Exp. 1 | Dependencies | Spanish | 0 | 16 | 30 | Average |
| Buerki 2019 Exp. 1 | Dependencies | French | 200 | 24 | 64 | Data |
| Buerki 2019 Exp. 2 | Dependencies | French | 0 | 38 | 16 | Data |
| Buerki 2019 Exp. 2 | Dependencies | French | 200 | 38 | 16 | Data |
| Buerki 2019 Exp. 3 | Dependencies | French | 0 | 27 | 48 | Data |
| Buerki 2019 Exp. 3 | Dependencies | French | 200 | 27 | 48 | Data |
| Buerki 2019 Exp. 4 | Dependencies | French | 0 | 30 | 48 | Data |
| Buerki 2019 Exp. 4 | Dependencies | French | 200 | 30 | 48 | Data |
| Alario 2002 Exp. 1 | Dependencies | French | 0 | 23 | 36 | Average |
| Buerki 2016 Exp. 1 | No dependencies | German | 0 | 20 | 70 | Data |

## Appendix B

*Meta-analytic estimate of Gender congruency effect in Germanic and Slavic languages at an SOA of 0 with different priors*

| Prior | Estimate | lower | upper | Tau | Tau.lower | Tau.upper |
|---|---|---|---|---|---|---|
| N(0,100) | 18.52 | 14.20 | 23.00 | 2.66 | 0.10 | 8.03 |
| N(0,200) | 18.59 | 14.23 | 23.09 | 2.76 | 0.10 | 8.37 |
| Uniform(-100,100) | 18.56 | 14.39 | 23.01 | 2.77 | 0.10 | 8.10 |

*Meta-analytic estimate of Gender congruency effect in Romance languages with phonological dependencies at an SOA of 0 with different priors*

| Prior | Estimate | lower | upper | tau | Tau.lower | Tau.upper |
|---|---|---|---|---|---|---|
| N(0,100) | 0.89 | -3.95 | 5.65 | 2.38 | 0.10 | 7.08 |
| N(0,200) | 0.95 | -3.70 | 5.59 | 2.43 | 0.10 | 7.29 |
| Uniform(-100,100) | 0.89 | -3.69 | 5.47 | 2.40 | 0.09 | 6.98 |

## Appendix C

*Meta-analytic estimate of Gender congruency effect in Romance languages with dependencies at late SOAs with different priors*

| Prior | Estimate | lower | upper | tau | Tau.lower | Tau.upper |
|---|---|---|---|---|---|---|
| N(0,100) | 8.08 | -0.22 | 17.72 | 8.98 | 0.43 | 21.93 |
| N(0,200) | 8.03 | -0.30 | 17.05 | 9.18 | 0.50 | 22.61 |
| Uniform(-100,100) | 8.11 | -0.19 | 17.32 | 9.07 | 0.46 | 22.47 |